\pdfoutput=1
\documentclass[11pt]{article}
\usepackage[]{SciBERT-CNN}
\usepackage{times}
\usepackage{latexsym}
\usepackage{graphicx}
\usepackage[T1]{fontenc}
\usepackage[utf8]{inputenc}
\usepackage{microtype}
\usepackage{inconsolata}
\usepackage{lineno}
\usepackage{booktabs}
\usepackage{array}
\usepackage{tabularx}
\usepackage{multicol}
\usepackage{hyperref}
\usepackage{float}
\usepackage{amsmath}

\title{Empowering Interdisciplinary Research with
BERT-Based Models:
\\An Approach Through SciBERT-CNN with Topic Modeling}

\author{Darya Likhareva, \ Hamsini Sankaran, \ Sivakumar Thiyagarajan \\
  University of California, Berkeley \\
  \texttt{\{likhareva,hamsini.sankaran,siva.thiyagarajan\}@berkeley.edu}}

\begin{document}
\maketitle
\begin{abstract}
Researchers must stay current in their fields by regularly reviewing academic literature, a task complicated by the daily publication of thousands of papers. Traditional multi-label text classification methods often ignore semantic relationships and fail to address the inherent class imbalances. This paper introduces a novel approach using the SciBERT model and CNNs to systematically categorize academic abstracts from the Elsevier OA CC-BY corpus. We use a multi-segment input strategy that processes abstracts, body text, titles, and keywords obtained via BERT topic modeling through SciBERT. Here, the [CLS] token embeddings capture the contextual representation of each segment, concatenated and processed through a CNN. The CNN uses convolution and pooling to enhance feature extraction and reduce dimensionality, optimizing the data for classification. Additionally, we incorporate class weights based on label frequency to address the class imbalance, significantly improving the classification F1 score and enhancing text classification systems and literature review efficiency. 

\end{abstract}


\section{Introduction}
According to Brown and Columbia University Revenue Statistics, 64 million academic papers have been published, with the rate of newly published articles steadily rising. In this sea of information, it is virtually impossible for researchers to manually review every manuscript \cite{wordsrated2023}.  In fact, 2021 experienced a surge in publishing, with a 7.62\% increase compared to the previous year. This increase not only overwhelms researchers but inhibits important breakthroughs due to poor or inadequate classification. There is a need for improved classification methods to ensure all research is recognized and utilized.

Multi-label text classification is a popular technique, offering extensive practical applications, particularly in tag recommendation, sentiment analysis, and information retrieval. Automated scientific text classification labeling is the first step in determining an academic paper's relevance. Existing models often fail to accurately classify academic papers due to their inability to understand the context and multidisciplinary nature of research. Our approach pulls from advancements in NLP to better capture and classify the nuanced and complex nature of academic texts. 

Existing BERT models, while powerful, often fall short in academic text classification due to their generalist nature and may not fully capture the extended context of the data. Our fine-tuned SciBERT-CNN topic model utilizes a combination of features (abstract, body text, title, and BERT topic keywords) to enhance classification accuracy and set a new benchmark for academic text classification. Our experiments show significant reductions in misclassification and improvements in accuracy and efficiency compared to a standard BERT model. Additionally, our model is specially tailored for the complex vocabulary and syntax of scientific literature, improving the interpretation and classification of interdisciplinary research papers.

For this work, we utilized the Elsevier OA CC-BY corpus dataset.\cite{kershaw2020} We strategically chose this data set as it contains a diverse and comprehensive range of open-access articles across 27 different disciplines, providing us with a robust foundation for training and evaluating our classification model. We pulled the dataset directly from HuggingFace \cite{huggingfacelink}.

This project tackles several challenges: the complexity and nuances of scientific language, class imbalance, and a large volume of text. 

\section{Background}
Recent research has shown that approaches in multi-label text classification have significantly enhanced efficiency and effectiveness.

In their research, Beltagy et al. introduced a new BERT-based model, SCIBERT, pre-trained on a large corpus of 1.14 million papers in scientific literature, mainly from the fields of computer science and biomedical sciences. The model has demonstrated significant improvements in performance across various scientific NLP tasks, surpassing BERT and achieving state-of-the-art results in several cases \cite{beltagy2019scibert}.  

The research by Chang et al. addresses the problems in ensembling BERT models requiring more computation by using a novel ensembling approach that uses multiple CLS tokens to enhance the efficiency and performance of BERT models. Multi-CLS BERT achieves near-parity with a 5-way BERT ensemble, offering significant gains with lower computational and memory costs. The research concluded that the Multi-CLS BERT approach is effective, especially with limited compute resources and BERT token limitation~\cite{chang2022multi}. 

Additionally, our dataset also has a class imbalance issue, and we consulted existing literature. The ‘Balancing Methods for Multi-label Text Classification with Long-Tailed Class Distribution’ paper introduces balancing loss functions in NLP to improve classification in class imbalance and label linkage. Experiments conducted on a general dataset (Reuters-21578) and a domain-specific dataset (PubMed) reveal that a distribution-balanced loss function outperforms traditional loss functions, addressing class imbalance and label dependencies~\cite{kershaw2020}.

\begin{figure}[htbp!]
\centering
\includegraphics[width=0.5\textwidth]{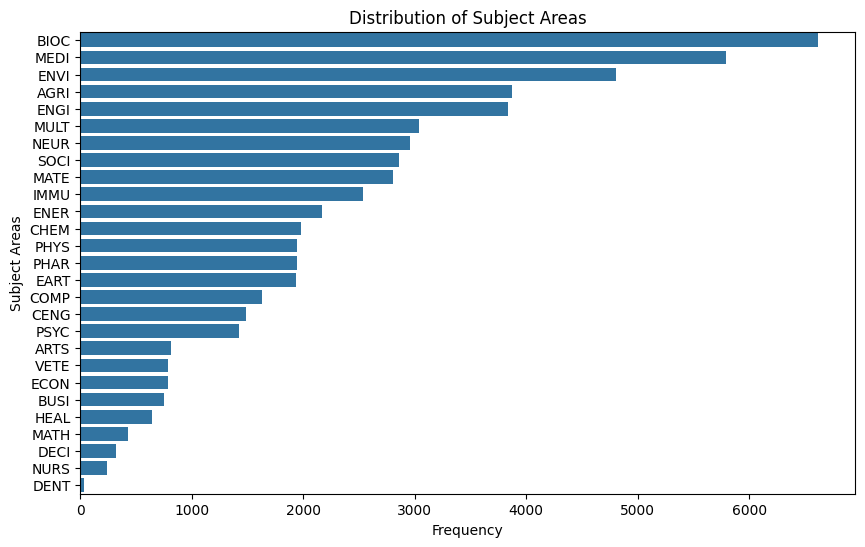}
\caption{The 27 Classes: Subject Area Imbalance}
\label{fig: Subject Area Imbalance}
\end{figure}

We also looked at a study based on the Label-Specific Attention Network (LSAN) to enhance multi-label text classification by utilizing document and label semantic information to craft label-specific document representations. LSAN effectively navigates the semantic connections between documents and labels by applying a self-attention mechanism and an adaptive fusion strategy. This was proven to work well on various datasets, showing that LSAN is reliable and can handle complicated tasks~\cite{huang2021}.

In our examination of recent methodologies for textual analysis within scientific publications, a notable approach from \cite{farber2021identifying} has been identified that leverages both SciBERT and Convolutional Neural Networks (CNNs) for the classification of method and dataset references. The cited paper presents a comparative study where the fine-tuned SciBERT model exhibits superior performance in recognizing 'METHOD' entities when provided with a single sentence. This model achieves impressive recall metrics, which is indicative of its ability to identify relevant entities within a limited context.
Building upon the referenced approach, Our approach employs a hybrid model that uses SciBERT, CNN, BERT topic keywords and a class weights strategy adjusting for label frequency in the training data to enhance multi-label text classification.

\begin{figure}[htbp!]
\centering
\includegraphics[width=0.5\textwidth]{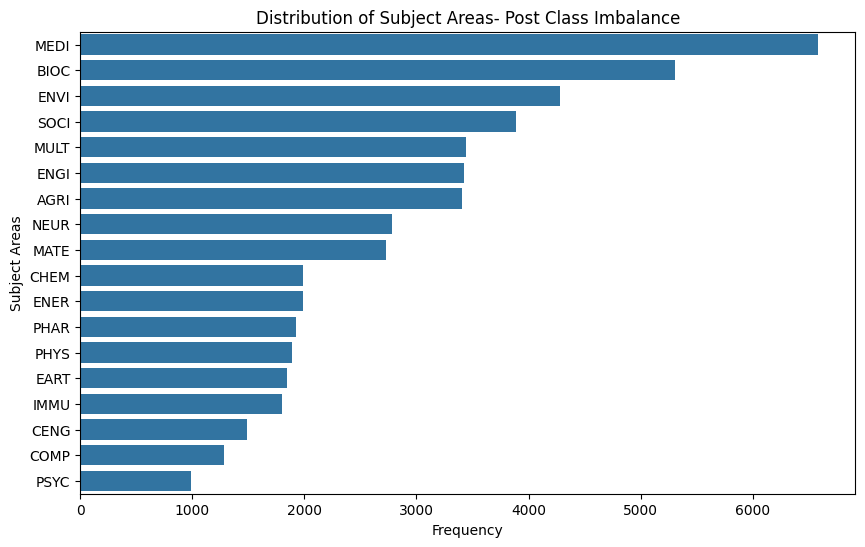}
\caption{Resulting 18 Class Subject Areas}
\label{fig:18 Class Subject Areas}
\end{figure}

\begin{figure*}[ht]
    \centering
    \includegraphics[width=1\textwidth]{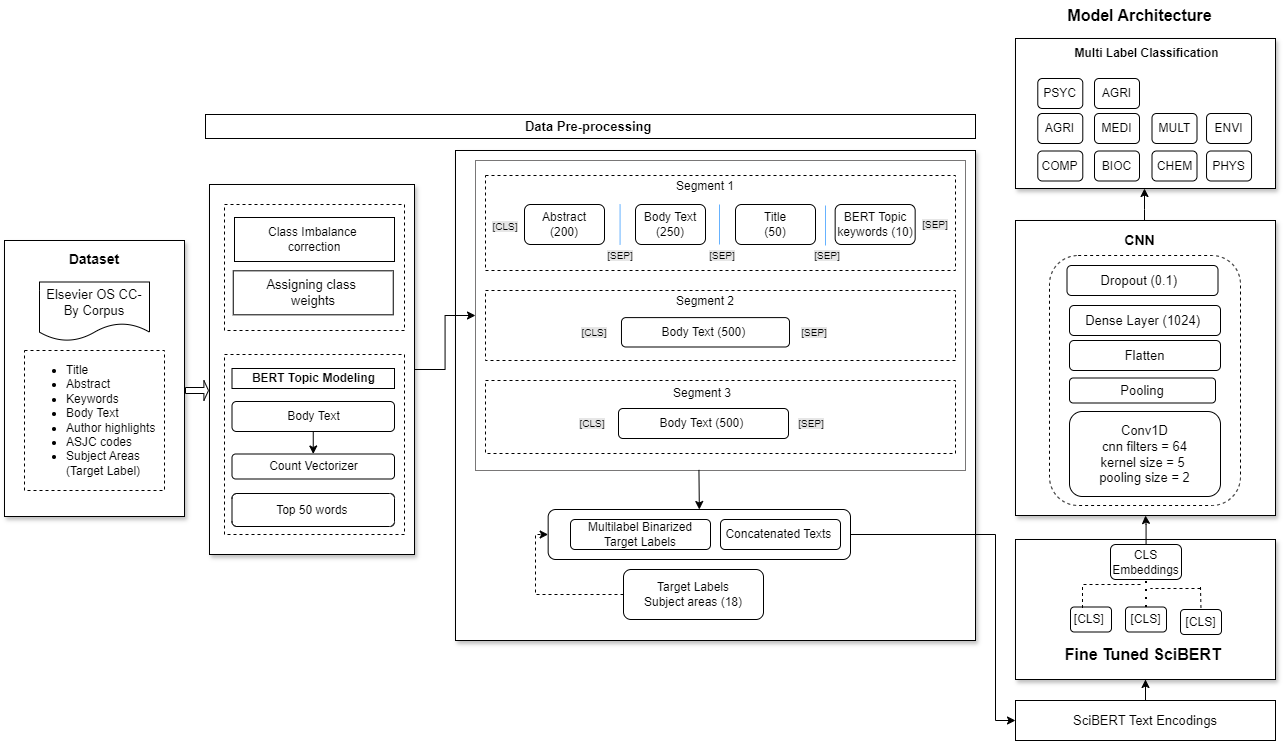}
    \caption{SciBERT-CNN Architecture}
    \label{fig:SciBERT-CNN Architecture}
\end{figure*}

\section{Methods}
\subsection{Baseline}
We chose the Elsevier OA CC-BY corpus dataset \cite{kershaw2020}, which consists of 40,000 open-access articles from across Elsevier's journals. We chose subject areas as the target labels and started with abstract as our input into the model. We established a BERT baseline model, pre-trained on large-scale general-domain corpora, and implemented a standard transformer-based architecture customized for multi-label text classification on all 27 classes. We saw significant variance across different subject areas, with certain categories like 'MULT' and 'NEUR' achieving relatively high F1 scores, while others such as 'DECI' and 'MATH' with very low F1 scores.

\subsection{Dataset: Class Imbalance}
As we experimented, we encountered significant class imbalance, as seen in Figure \ref{fig: Subject Area Imbalance}. To mitigate bias towards frequent labels, we implemented several measures to balance our dataset.

We started with consolidating the original 27 classes into 18 by grouping smaller frequency, poorly performing areas into broader subject categories: 'VETE', 'HEAL', 'DENT', and 'NURS' were reassigned to 'MEDI' under medical; 'ECON', 'ARTS', 'BUSI', and 'DECI' were combined into 'SOCI' within science and humanities; and 'MATH' was merged into 'MULT' for multidisciplinary studies, as seen in Figure \ref{fig:18 Class Subject Areas}. Through experiments on our baseline BERT model, we saw noted improvement in individual F1 scores; more importantly, we no longer had any categories with a zero weighted F1 score.

Continuing to address our class imbalance, we implemented a weighted average function that assigned higher weights to rarer labels. As referenced by \cite{huang2021}, distribution-based loss functions address the class imbalance and label linkage problems, and they outperform commonly used loss functions. Less frequent labels—which are typically underrepresented and can be overshadowed by more common ones—receive higher weights. We ensured that the weights were scaled and did not disproportionately affect the learning process and normalized their sum to equal one. This normalization balances the impact of each label's weight during model training, heightening the model's sensitivity to less common labels. We did this to ensure our model does not bias towards samples with more frequent labels, and through this implementation, we saw further improvements from the baseline. 

\subsection{Inputs and Parameters in the Model}

We experimented with various combinations of inputs in our model, as shown in Table \ref{tab:evaluation_results}. Through our experimentation, we determined that the optimal input combination for our model was the abstract, body text, title, and the top 10 important words from the body text generated from BERT topic modeling. Outlined below is a comprehensive breakdown of our data processing methodology:

\textbf{Data Segmentation and Preparation:} We divide the text data into segments of abstracts, titles, and body texts and prepare them by trimming each to specific word limits to ensure conciseness and relevance. We initially divide each document into an abstract (limited to 200 words), a title (capped at 50 words), and body text (250 words). Additionally, we enrich this segment by identifying and including the top 10 keywords from the body text using BERT-based topic modeling. We then extended the body text to two more segments by adding an additional 500 words each segment to deepen the analysis. 

\textbf{Topic Modeling:} We used a BERT-based topic modeling technique on the body text to distill topics that encapsulate the corpus's semantic content, aiding in dimensionality reduction and contextual understanding. We then concatenated the abstract, title, and body text and extracted topics to the model's input.

\textbf{Multi-Label Binarization and Tokenization:} Subject area labels are encoded using MultiLabelBinarizer to prepare for multi-label classification, converting categorical labels into a binary format suitable for the model to process. The concatenated texts are tokenized using the SciBERT tokenizer, creating the necessary inputs for the model, including input IDs and attention masks.

\textbf{Class Weights:} In our model, we addressed the challenge of long-tailed class imbalance by adjusting the class weights. We computed the class weights based on the frequency of each label within the training data, allowing us to assign a higher weight to less frequent labels and mitigate the dominance of more common ones as indicated in equations (\ref{eq:eq1}), 
 (\ref{eq:eq2}) and (\ref{eq:eq3}). This adjustment helps to enhance the model's sensitivity to underrepresented classes. Here is a mathematical representation of the method used to calculate the class weights:
Let L be the set of labels in the dataset, and let freq(\textit{l}) denote the frequency of label l in L. The weight \textit{w}(\textit{l}) assigned to each label \textit{l} is computed as follows:

\begin{equation}
w(l) = \frac{1}{\text{freq}(l)}
\label{eq:eq1}
\end{equation}

To normalize these weights so that they sum to 1, we apply:

\begin{equation}
w(l) = \frac{w(l)}{\sum_{l \in L} w(l)}
\label{eq:eq2}
\end{equation}

For each sample \( s \) with labels \( l_s \subset L \), the sample weight \( w_s \) is the sum of the weights of its labels:

\begin{equation}
w_s = \sum_{l \in l_s} w(l)
\label{eq:eq3}
\end{equation}

\subsection{Model Architecture}
Our approach employs a hybrid neural architecture that combines SciBERT's contextual analysis with CNN's pattern recognition to classify scientific texts into 18 subject areas. By fine-tuning SciBERT on a scientific corpus and applying CNNs to the derived embeddings, we enhance our model's ability to discern and categorize nuanced scientific topics effectively. The framework is presented in Figure \ref{fig:SciBERT-CNN Architecture}.

We experimented with various transformer-based architectures to identify the most effective model for multi-label classification tasks. As mentioned, we started with a foundational BERT model and felt that we could explore other model architectures that better suited our academic domain. Our next step was to leverage RoBERTa, an optimized variant of BERT, known for its improved training regimen and more robust results. We also evaluated using Longformer, which offered significant improvements in handling longer data lengths, and we felt that this would be useful for our dataset due the the long body text feature.  

\begin{table*}[htbp] 
\caption{Model Evaluation Results (Weighted Average: Precision, Recall, F1)}
\label{tab:evaluation_results}
\scriptsize
\begin{tabularx}{\textwidth}{@{}X X X c c c@{}}
\toprule
\textbf{Model} & \textbf{Embeddings} & \textbf{Feature Modifications} & \textbf{Pr} & \textbf{Recall} & \textbf{F1} \\
\midrule
BERT [Baseline] & BERT & abs only & 0.76 & 0.51 & 0.59 \\
\midrule
RoBERTa & RoBERTa & abs only & 0.73 & 0.54 & 0.60 \\
\midrule
Longformer & Longformer & abs only & 0.70 & 0.62 & 0.63 \\
\midrule
SciBERT & SciBERT & abs only & 0.73 & 0.59 & 0.64 \\
\midrule
SciBERT & SciBERT & abs + body\_text & 0.70 & 0.68 & 0.68 \\
\midrule
SciBERT & SciBERT & abs + body\_text, CLS embeddings & 0.69 & 0.71 & 0.69 \\
\midrule
SciBERT & SciBERT & abs + body\_text + title + keywords, CLS embeddings + CNN & 0.73 & 0.69 & 0.70 \\
\midrule
SciBERT & SciBERT & abs + body\_text + title + top 10 important words from body\_text, Keybert & 0.69 & 0.71 & 0.69 \\
\midrule
SciBERT [Final Model] & SciBERT & abs + body\_text + title + top 10 important words from body\_text, BERT topic modeling & 0.69 & 0.72 & 0.70 \\
\bottomrule
\end{tabularx}
\end{table*}

We looked into domain-specific models. We ultimately choose SciBERT as our model architecture of choice, which aligns closely with the academic nature of the Elsevier corpus \cite{kershaw2020}. We enhanced model performance through a series of experiments. 

Out of all the experiments conducted, integrating SciBERT with a Convolutional Neural Network (CNN) and BERT topic modeling yielded the most favorable results. This combination proved to be the superior approach and boosted the performance of text classification. Our architecture utilizes SciBERT to capture contextual embeddings from the tokenized text. The [CLS] token embeddings from SciBERT serve as a summary of the document's context and are fed into a CNN structure, which includes a convolutional layer followed by a max-pooling layer, while [SEP] tokens are appended to indicate segment separations. This structure is specifically designed to capture and distill local contextual features from the embeddings. A dropout strategy is incorporated to prevent overfitting, ensuring that the model generalizes well to new, unseen data. The outputs from the CNN are passed through a dense layer to aid in high-level feature detection before being subjected to the final classification layer. The classification layer employs a sigmoid activation function to output probabilities for each of the target classes, enabling the model to perform multi-label classification. We further fine-tuned the SciBERT to align the model with the domain-specific characteristics of the Elsevier corpus. This process was essential to tailor the pre-trained model to the particular lexical and semantic features of the scientific text, optimizing performance for the classification task.

The intuition behind our proposed approach stems from the understanding that scientific texts are dense with specialized language and complex concepts that can span across multiple disciplines. SciBERT is able to capture contextual embeddings. However, while SciBERT captures the global context of text, it may not focus on local patterns. Hence, we used CNN, which is excellent at detecting patterns within local regions of data.

We chose to measure success by evaluating both the weighted average F1 scores as well as the individual F1 scores for each class. The F1 score helps measure how well our model accurately identifies all relevant labels without penalizing for false positives and false negatives, making it ideal for assessing performance across our very diverse and class-imbalanced Elsevier corpus. For clarity, we include the weighted average F1 score across all our experiments as a point of comparison Table \ref{table:model_performance}, and we also show a breakdown of the individual F1 scores of our baseline model compared to our best SciBERT/CNN model. 

\section{Results and Discussion}
Confusion matrices were generated for each of the subject area labels, serving as a detailed evaluative measure and ROC curves were also used as additional metrics to further gauge the success of our model. This enabled us to perform a nuanced analysis of model performance across all labels. We compiled a detailed classification report encompassing each label's precision, recall, and F1 score to supplement our understanding of the model's predictive power. This multifaceted approach to performance evaluation illuminated the strengths and weaknesses of our models on a label-by-label basis, allowing us to discern their efficacy in the multi-label text classification context comprehensively.

\begin{table}[h]
\centering
\caption{Comparison of Baseline and Best Model Performance}
\label{table:model_performance}
\renewcommand{\arraystretch}{1.5}
\resizebox{\columnwidth}{!}{
\small
\begin{tabular}{|l|c|c|c|c|c|}
\hline
\textbf{Label} & \textbf{Baseline F1-Score} & \multicolumn{4}{c|}{\textbf{Best Model}} \\ \cline{3-6}
 & & \textbf{Precision} & \textbf{Recall} & \textbf{F1-Score} & \textbf{Support} \\ \hline
AGRI & 0.64 & 0.71 & 0.78 & 0.74 & 413 \\ \hline
BIOC & 0.34 & 0.69 & 0.69 & 0.69 & 653 \\ \hline
CENG & 0.19 & 0.41 & 0.68 & 0.51 & 189 \\ \hline
CHEM & 0.31 & 0.55 & 0.66 & 0.60 & 248 \\ \hline
COMP & 0.41 & 0.47 & 0.57 & 0.51 & 157 \\ \hline
EART & 0.71 & 0.57 & 0.93 & 0.71 & 217 \\ \hline
ENER & 0.65 & 0.64 & 0.63 & 0.64 & 235 \\ \hline
ENGI & 0.58 & 0.56 & 0.76 & 0.64 & 444 \\ \hline
ENVI & 0.67 & 0.67 & 0.60 & 0.63 & 512 \\ \hline
IMMU & 0.62 & 0.67 & 0.59 & 0.63 & 238 \\ \hline
MATE & 0.65 & 0.70 & 0.89 & 0.78 & 331 \\ \hline
MEDI & 0.71 & 0.82 & 0.71 & 0.76 & 851 \\ \hline
MULT & 0.79 & 0.98 & 0.68 & 0.80 & 467 \\ \hline
NEUR & 0.81 & 0.86 & 0.86 & 0.86 & 394 \\ \hline
PHAR & 0.47 & 0.51 & 0.75 & 0.61 & 219 \\ \hline
PHYS & 0.38 & 0.49 & 0.75 & 0.59 & 203 \\ \hline
PSYC & 0.50 & 0.46 & 0.81 & 0.58 & 120 \\ \hline
SOCI & 0.52 & 0.87 & 0.69 & 0.77 & 470 \\ \hline
\multicolumn{1}{|r|}{micro avg} & 0.61 & 0.67 & 0.72 & 0.69 & 6361 \\ \hline
\multicolumn{1}{|r|}{macro avg} & 0.46 & 0.65 & 0.72 & 0.67 & 6361 \\ \hline
\multicolumn{1}{|r|}{weighted avg} & 0.59 & 0.70 & 0.72 & 0.70 & 6361 \\ \hline
\multicolumn{1}{|r|}{samples avg} & 0.60 & 0.72 & 0.74 & 0.70 & 6361 \\ \hline
\end{tabular}
}
\end{table}

\begin{figure}
    \centering
    \includegraphics[width=1\linewidth]{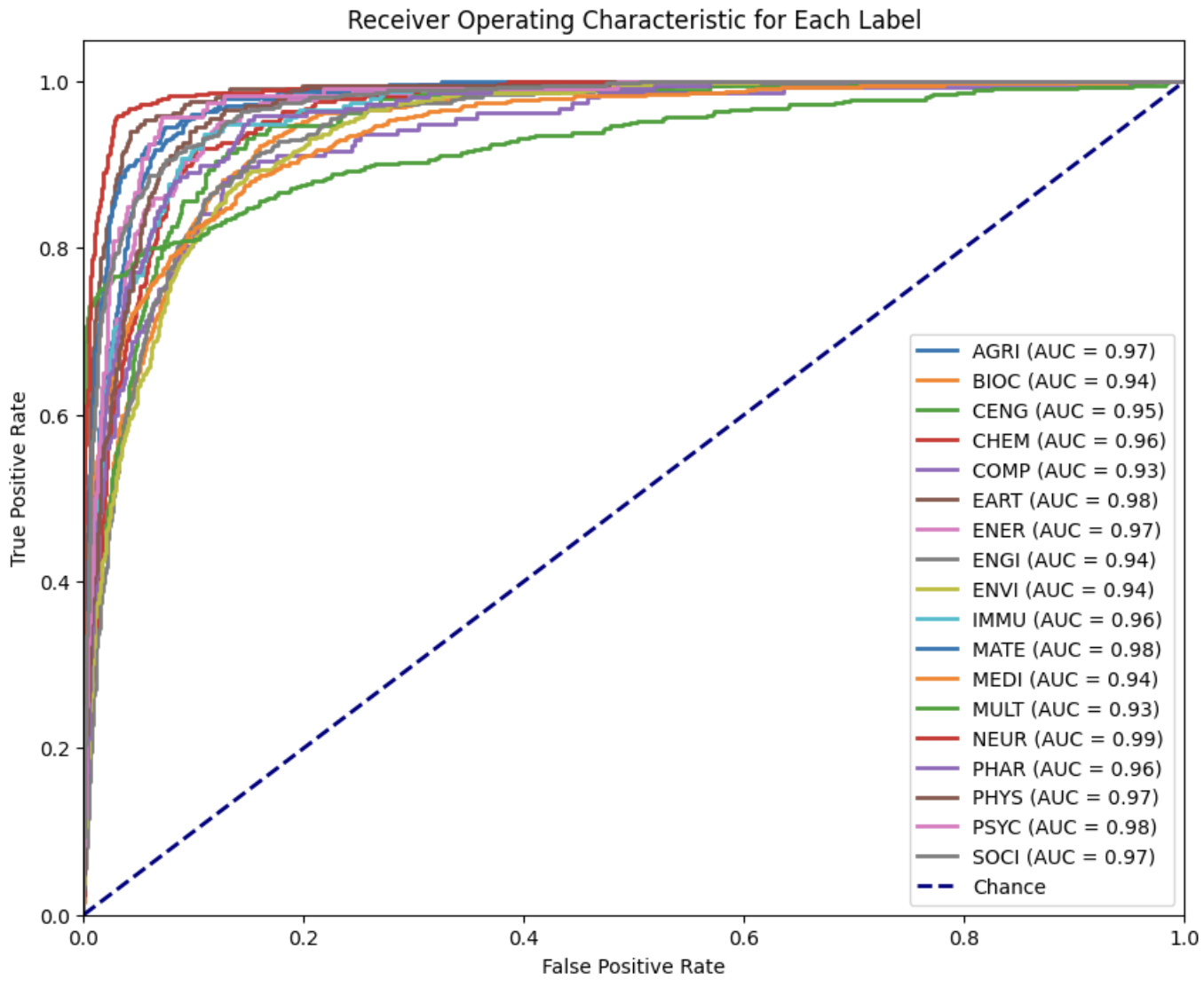}
    \caption{Receiver Operating Characteristics of each label}
    \label{fig:receiver_operating}
\end{figure}

As seen in Table \ref{table:model_performance}, we saw significant improvement in the performance of our SciBERT/CNN model compared to the baseline BERT model across almost all categories. For instance, in categories like ‘BIOL,’ the F1 score increased from 0.34 to 0.72, and in ‘CENG,’ it improved from 0.19 to 0.70. Our fine-tuning implemented in the SciBERT-CNN model, through the CNN's convolutional and max-pooling layers, dropout strategy, dense layer, and classification layer with a sigmoid function, have effectively addressed key weaknesses of the baseline model. Some classes like ‘DENT’, ‘EART’, and others were moved to a larger class to address issues like class imbalance and improve overall model robustness and accuracy. The ROC curve chart, as shown in Figure \ref{fig:receiver_operating}, demonstrates excellent classifier performance across multiple categories, with AUC values ranging from 0.93 to 0.99, indicating strong separability between positive and negative cases. The best-performing categories are 'EART' (Earth) and 'MATE' (Materials). Overall, the model is consistent and reliable in its predictions across the various labels.

\begin{figure}[htbp!]
\centering
\includegraphics[width=\linewidth]{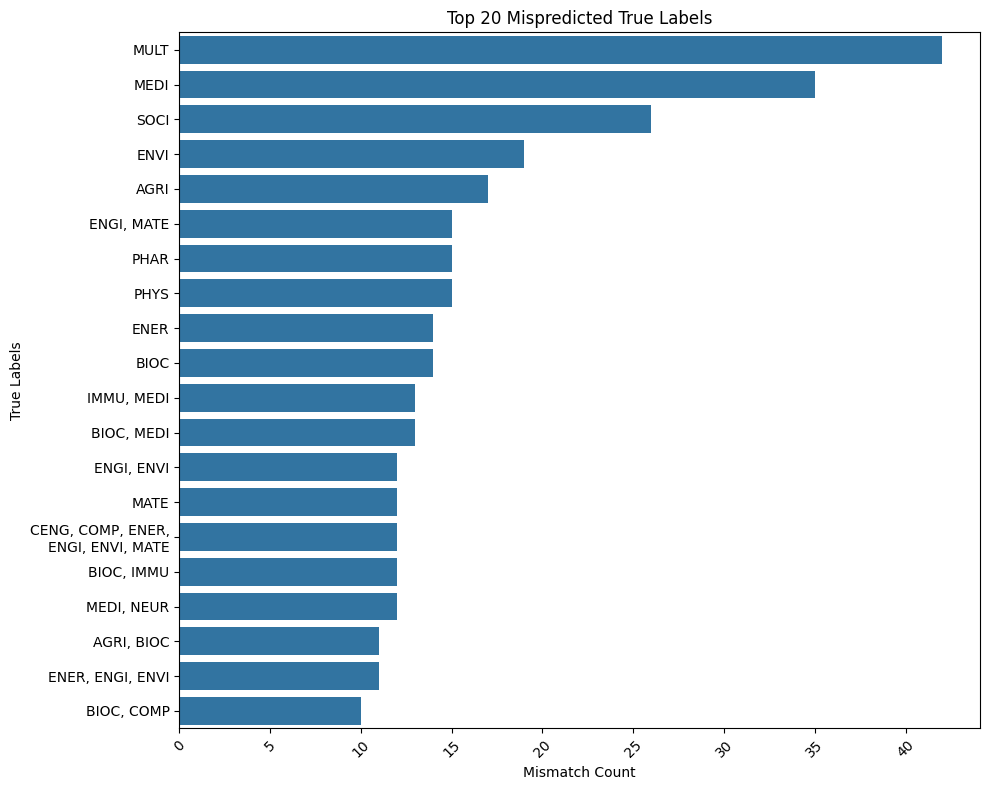}
\caption{Misplaced labels}
\label{fig:Misplaced labels}
\end{figure}

\subsection{Error Analysis} 
We further analyzed the most common misclassifications, as seen in Figure \ref{fig:Misplaced labels}. The ‘MULT’ label has the highest number of mispredictions. This outcome is expected given the label’s broad definition, which spans a diverse range of subject areas. Additionally, the ‘MEDI’ category, while it has a high frequency in our dataset, the medical field is known for its complex and specialized vocabulary. Medical terminology often overlaps with other disciplines, such as biology (‘BIOC’), which can lead to further misclassifications. More so, most cases of misclassifications resulted in a no-label prediction, indicating that the model does not receive enough context to make accurate predictions. As noted in our conclusion, future work in enhancing the data set with more contextual features and augmenting the data set could improve the model's ability to correctly classify these categories.

\textbf{Class Imbalance:} Initially, some categories like ‘DENT’ and ‘NURS’ had very low F1 scores (0.0 and 0.05, respectively) due to low frequency in the dataset. Merging these with larger classes helped in sharing characteristics and improving the model's ability to learn from more examples.

\textbf{Shortcomings of SciBERT:} SciBERT is trained on a corpus of scientific literature mainly from computer science and biomedical fields and might face limitations when applied to datasets containing humanities content, such as the Elsevier corpus. While the structure of academic articles is similar from article to article, linguistic patterns, terminology, and content structure in humanities articles differ significantly from those in scientific and technical articles. This misalignment can lead to classification errors, impacting the model’s performance and accuracy in handling humanities content within our dataset.

\textbf{Structural Issues of Journal Classifications:}
Since subject labels (and broader ASJC codes) for individual academic articles are assigned to the broader categories of the journals in which they are published, the subject labels assigned are not always a perfect fit for the context of the actual paper. We suspect that some of our misclassifications are due to overly broad categorizations, where the specific content of an article might not align well with the general scope of the journal. As a result, our model might struggle with accuracy in cases where the article's content targets a very specific subject area but is assigned a much broader label. This highlights yet another challenge of multilabel classification in journal articles.

\textbf{Shortage of Computational Resources:} 
We faced constraints due to the computational demands of more advanced models and in some instances, observed a high input of compute power for relatively low improvements in performance. 

\section{Conclusion}
The primary challenge we addressed was the multi-label text classification of scientific articles within the Elsevier corpus. This was a complex issue, as we had data complexities we needed to solve in order to accurately categorize documents into multiple overlapping academic subject areas. 

Our model for multi-label text classification of the Elsevier corpus demonstrated notable performance gains by integrating SciBERT with BERT topic modeling. By focusing on abstracts, body text segments, titles, and a selection of the top 10 words, we surpassed other BERT-based model benchmarks. 
The hybrid combination of SciBERT and CNN significantly improved individual and weighted F1 scores. 

Future efforts will explore data augmentation and the integration of domain-specific keywords to bolster underrepresented classes and refine overall label performance. Additionally, we aim to develop a concise yet effective representation of body texts that emphasizes key terms to further enhance model input. With this, we anticipate even greater performance gains in more accurate classification of scientific text labels. 

\bibliographystyle{apalike}  
\bibliography{SciBERT-CNN}

\begin{thebibliography}{}

\bibitem[Beltagy et~al., 2019]{beltagy2019scibert}
Beltagy, I., Lo, K., and Cohan, A. (2019).
\newblock Scibert: A pretrained language model for scientific text.
\newblock {\em arXiv preprint arXiv:1903.10676}.

\bibitem[Brousse, 2022]{huggingfacelink}
Brousse, N. (2022).
\newblock orieg/elsevier-oa-cc-by from hugging face.
\newblock Available at: \url{https://huggingface.co/datasets/orieg/elsevier-oa-cc-by}.

\bibitem[Chang et~al., 2022]{chang2022multi}
Chang, H.-S., Sun, R.-Y., Ricci, K., and McCallum, A. (2022).
\newblock Multi-cls bert: An efficient alternative to traditional ensembling.
\newblock {\em arXiv preprint arXiv:2210.05043}.

\bibitem[Chen et~al., 2023]{chen2023}
Chen, X., Yin, Y., and Feng, T. (2023).
\newblock Multi-label text classification based on bert and label attention mechanism.
\newblock In {\em 2023 Asia-Pacific Conference on Image Processing, Electronics and Computers (IPEC)}, pages 386--390. IEEE.

\bibitem[Curcic, 2023]{wordsrated2023}
Curcic, D. (2023).
\newblock Number of academic papers published per year –wordsrated.
\newblock Available at: \url{https://wordsrated.com/number-of-academic-papers-published-per-year/}.

\bibitem[F{\"a}rber et~al., 2021]{farber2021identifying}
F{\"a}rber, M., Albers, A., and Sch{\"u}ber, F. (2021).
\newblock Identifying used methods and datasets in scientific publications.
\newblock In {\em SDU@ AAAI}.

\bibitem[Huang et~al., 2021]{huang2021}
Huang, Y., Giledereli, B., K{\"o}ksal, A., {\"O}zg{\"u}r, A., and Ozkirimli, E. (2021).
\newblock Balancing methods for multi-label text classification with long-tailed class distribution.
\newblock {\em arXiv preprint arXiv:2109.04712}.

\bibitem[Kershaw and Koeling, 2020]{kershaw2020}
Kershaw, D. and Koeling, R. (2020).
\newblock Elsevier oa cc-by corpus.
\newblock {\em arXiv preprint arXiv:2008.00774}.

\end{thebibliography}

\clearpage
\appendix
\section{Appendix}
\subsection{More info about the Dataset}
\begin{figure}[htbp!]
\centering
\includegraphics[width=\linewidth]{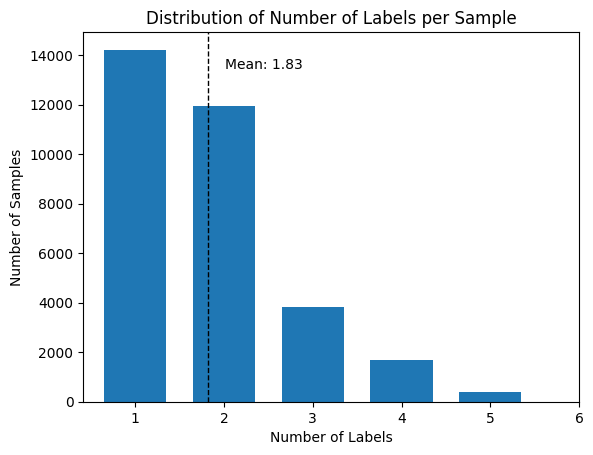}
\caption{Distribution of Subject area labels in the dataset}
\label{fig:Misplaced_labels}
\end{figure}

The Elsevier OA CC-BY corpus dataset \cite{wordsrated2023} consists of 40,000 open-access articles from across Elsevier's journals, representing a diverse research discipline (Table \ref{table:articles_by_discipline}), annotated with several metadata about the articles, including ASJC Codes and subject areas. Figure \ref{fig:Misplaced_labels} shows the distribution of number of subject area labels in the dataset.
\begin{table}[h]
\centering
\footnotesize 
\caption{Number of Articles by Discipline}
\label{table:articles_by_discipline}
\begin{tabular}{|l|l|r|}
\hline
\textbf{Label} & \textbf{Discipline} & \textbf{No. of Articles} \\ \hline
MULT & General & 310 \\ \hline
AGRI & Agricultural Sciences & 3985 \\ \hline
ARTS & Arts & 1014 \\ \hline
BIOC & Biochemistry & 8417 \\ \hline
BUSI & Business & 1002 \\ \hline
CENG & Chemical Engineering & 2196 \\ \hline
CHEM & Chemistry & 2749 \\ \hline
COMP & Computer Science & 3004 \\ \hline
DECI & Decision Sciences & 530 \\ \hline
EART & Earth Sciences & 2764 \\ \hline
ECON & Economics & 1081 \\ \hline
ENER & Energy & 2845 \\ \hline
ENGR & Engineering & 5962 \\ \hline
ENVI & Environmental Science & 6241 \\ \hline
IMMU & Immunology & 3258 \\ \hline
MATE & Materials Science & 4008 \\ \hline
MATH & Mathematics & 1561 \\ \hline
MEDI & Medicine & 9225 \\ \hline
NEUR & Neuroscience & 3277 \\ \hline
NURS & Nursing & 310 \\ \hline
PHAR & Pharmacology & 2233 \\ \hline
PHYS & Physics and Astronomy & 3927 \\ \hline
PSYC & Psychology & 1796 \\ \hline
SOCI & Social Sciences & 3623 \\ \hline
VETE & Veterinary & 1010 \\ \hline
DENT & Dentistry & 43 \\ \hline
HEAL & Health Professions & 774 \\ \hline
\end{tabular}
\end{table}

\subsection{BERT Label Attention Mechanism (LAM)} 
Our team also implemented a novel Label Attention Mechanism (LAM) layer on top of BERT, inspired by existing research \cite{chen2023}. The results of BERT's label attention mechanism are shown in Table \ref{table:bert_lam}. This mechanism was designed to fine-tune the influence of label semantics, yielding a drastic increase in both individual and overall F1 scores from the baseline BERT model. However, while LAM significantly enhanced BERT's performance, the same improvements were not observed when it was applied in conjunction with SciBERT. Despite our anticipation that LAM would provide similar benefits to SciBERT, the results indicated no major improvement in F1 scores, suggesting that the mechanism's effectiveness may be model-specific.

\begin{table}[h]
\centering
\caption{BERT LAM Performance}
\label{table:bert_lam}
\renewcommand{\arraystretch}{1.5}
\resizebox{\columnwidth}{!}{%
\begin{tabular}{lcccc}
\hline
\textbf{Label} & \textbf{Precision} & \textbf{Recall} & \textbf{F1-Score} & \textbf{Support} \\ 
\hline
AGRI & 0.73& 0.74& 0.73& 410\\ 
BIOC & 0.66& 0.77& 0.71& 630\\ 
CENG & 0.50& 0.32& 0.39& 191\\ 
CHEM & 0.56& 0.50& 0.53& 258\\ 
COMP & 0.75& 0.34& 0.47 & 148\\ 
EART & 0.87& 0.59& 0.70& 219\\ 
ENER & 0.73& 0.66& 0.69& 268\\ 
ENGI & 0.66& 0.54& 0.59& 425\\ 
ENVI & 0.69& 0.64& 0.67& 546\\ 
IMMU & 0.74& 0.49& 0.59& 228\\ 
MATE & 0.80& 0.71& 0.75& 337\\ 
MEDI & 0.84& 0.61& 0.71& 627\\ 
MULT & 0.85& 0.75& 0.79& 393\\ 
NEUR & 0.86& 0.73& 0.79& 383\\ 
PHAR & 0.75& 0.36& 0.49& 244\\ 
PHYS & 0.85& 0.24& 0.38& 227\\ 
PSYC & 0.79& 0.20& 0.32& 132\\ 
SOCI & 0.59& 0.78& 0.67& 209\\ 
\hline
\textbf{micro avg} & 0.73& 0.60& 0.66& 5875\\ 
\textbf{macro avg} & 0.73& 0.55& 0.61& 5875\\ 
\textbf{weighted avg} & 0.74& 0.60& 0.65& 5875\\ 
\textbf{samples avg} & 0.76& 0.66& 0.67& 5875\\ 
\hline
\end{tabular}
}
\end{table}

\subsection{SciBERT-CNN with KeyBERT }
In one of our experiments, we employed KeyBERT to identify the top 10 keywords from the 'body text' column of our datasets.  The results of KeyBERT are shown in Table \ref{table:keybert_performance}. A new column featuring these keywords was added to construct input sequences for our model, aimed at enhancing contextual comprehension. Although KeyBERT effectively identified pertinent terms, evidenced by F1 scores  above 50 for all labels, it significantly increased computational load and runtime. Due to these constraints, particularly the excessive GPU usage, we chose not to include KeyBERT in our final model. The experiment with KeyBERT provided valuable insights but was ultimately set aside in favor of more efficient methodologies.
\begin{table}[h]
\centering
\caption{SciBERT-CNN WITH KeyBERT}
\label{table:keybert_performance}
\renewcommand{\arraystretch}{1.5}
\resizebox{\columnwidth}{!}{%
\begin{tabular}{lcccc}
\hline
\textbf{Label} & \multicolumn{4}{c}{\textbf{Best Model}} \\ \cline{2-5} 
& \textbf{Precision} & \textbf{Recall} & \textbf{F1-Score} & \textbf{Support} \\ 
\hline
AGRI & 0.70& 0.83& 0.76& 413 \\ 
BIOC & 0.72& 0.69 & 0.70& 653 \\ 
CENG & 0.46& 0.62& 0.53& 189 \\ 
CHEM & 0.63& 0.54& 0.58& 248 \\ 
COMP &  0.57& 0.46& 0.51 & 157 \\ 
EART & 0.66& 0.79& 0.72& 217 \\ 
ENER & 0.60& 0.78& 0.68& 235 \\ 
ENGI & 0.66& 0.64& 0.65& 444 \\ 
ENVI & 0.59& 0.70& 0.64& 512 \\ 
IMMU & 0.53& 0.77& 0.63 & 238 \\ 
MATE & 0.83& 0.67& 0.74& 331 \\ 
MEDI & 0.79& 0.76& 0.77& 851 \\ 
MULT & 0.97& 0.60& 0.74& 467 \\ 
NEUR & 0.81& 0.92& 0.86 & 394 \\ 
PHAR & 0.61& 0.62& 0.61 & 219 \\ 
PHYS & 0.58& 0.45& 0.51& 203 \\ 
PSYC & 0.37& 0.88& 0.52& 120 \\ 
SOCI & 0.73& 0.75& 0.74& 470 \\ 
\hline
\textbf{micro avg} & 0.68& 0.71& 0.69& 6361 \\ 
\textbf{macro avg} & 0.66& 0.69& 0.66& 6361 \\ 
\textbf{weighted avg} & 0.70 & 0.71& 0.69& 6361 \\ 
\textbf{samples avg} & 0.72 & 0.74& 0.69& 6361 \\ 
\hline
\end{tabular}%
}
\end{table}

\begin{table}[h]
\centering
\caption{Unfreezing First 12 Layers of SciBERT}
\label{table:unfreezing_performance}
\renewcommand{\arraystretch}{1.5}
\resizebox{\columnwidth}{!}{%
\begin{tabular}{lcccc}
\hline
\textbf{Label} & \multicolumn{4}{c}{\textbf{Unfreezing 12 layers}} \\ \cline{2-5} 
& \textbf{Precision} & \textbf{Recall} & \textbf{F1-Score} & \textbf{Support} \\ 
\hline
AGRI & 0.68& 0.57& 0.62& 442\\ 
BIOC & 0.65& 0.65& 0.65& 665\\ 
CENG & 0.44& 0.31& 0.36& 184\\ 
CHEM & 0.53& 0.48& 0.51& 238\\ 
COMP & 0.48& 0.36& 0.41& 178\\ 
EART & 0.68& 0.52& 0.59& 233\\ 
ENER & 0.64& 0.52& 0.57& 260\\ 
ENGI & 0.61& 0.53& 0.57& 444 \\ 
ENVI & 0.52& 0.71& 0.60& 550\\ 
IMMU & 0.47& 0.61& 0.53& 223\\ 
MATE & 0.66& 0.72& 0.69& 345\\ 
MEDI & 0.78& 0.65& 0.71& 799\\ 
MULT & 0.98& 0.44& 0.60& 410\\ 
NEUR & 0.75& 0.69& 0.72& 308\\ 
PHAR & 0.53& 0.48& 0.50& 227\\ 
PHYS & 0.55& 0.52& 0.53& 217\\ 
PSYC & 0.46& 0.44& 0.45& 131\\ 
SOCI & 0.81& 0.59& 0.68& 520\\ 
\hline
\textbf{micro avg} & 0.64& 0.58& 0.61& 6361 \\ 
\textbf{macro avg} & 0.62& 0.54& 0.57& 6361 \\ 
\textbf{weighted avg} & 0.66& 0.58& 0.61& 6361 \\ 
\textbf{samples avg} & 0.71& 0.61& 0.58& 6361 \\ 
\hline
\end{tabular}%
}
\end{table}

In our pursuit to refine our SciBERT-based classification model, we experimented with unfreezing the first 12 layers of the pre-trained model. The results of unfreezing 12 layers  are shown in Table \ref{table:unfreezing_performance}. The rationale behind unfreezing these layers was to enable the adaptation of the model's lower-level representations to the nuances of our specific dataset, which often leads to a more tailored and hence more accurate model. Unfreezing layers in a deep learning model allows those layers to update their weights during training. However, in our case, this approach did not yield the anticipated improvement. Contrarily, it led to a reduced F1 score for the underrepresented classes. A potential reason for this could be that by unfreezing the first 12 layers, we allowed the model to overfit to certain aspects of the training data, leading to a loss of generalization ability.

\end{document}